\documentclass{llncs}
\usepackage{makeidx}  
\usepackage{url}      
\usepackage{subfigure}
\usepackage{amsfonts}
\usepackage{amsmath}
\usepackage{cancel}
\usepackage{graphicx}
\usepackage{color}
\usepackage{listings}
\usepackage{hyperref}
\hypersetup{
    colorlinks=true,
    linkcolor=blue,
    filecolor=magenta,      
    urlcolor=blue,
}
 
\usepackage{tikz}
\usepackage{ifthen}
\usepackage{xcolor}
\usepackage{multirow}
\usetikzlibrary{shadows.blur}
\usepackage{booktabs}
\usepackage{xspace}

\newlength{\forkmeoffset}
\definecolor{forkmebg}{HTML}{CC0000}
\definecolor{forkmefg}{HTML}{EEEEEE}

\newcommand{\etal}{\emph{et al.}\@\xspace}
\newcommand*{\eg}{e.g.\@\xspace}

\newcommand*{\cf}{c.f.\@\xspace}

\newcommand\blfootnote[1]{%
  \begingroup
  \renewcommand\thefootnote{}\footnote{#1}%
  \addtocounter{footnote}{-1}%
  \endgroup
}

\newcommand{\forkme}[1][west]{
	\ifthenelse{\equal{#1}{east}}{%
		\tikzset{forkmerot/.style={rotate=-45}}
	}{%
		\tikzset{forkmerot/.style={rotate=45}}
	}
	\begin{tikzpicture}[remember picture, overlay]
	\node[forkmerot, shift={(0, -\forkmeoffset)}] at (current page.north #1) {
		\begin{tikzpicture}[remember picture, overlay]
		\node[fill=forkmebg, text centered, minimum width=50em, minimum height=3.0em, blur shadow, shadow yshift=0pt, shadow xshift=0pt, shadow blur radius=.4em, shadow opacity=50, text=forkmefg](fmogh) at (0pt, 0pt) {   \fontfamily{phv}\selectfont\bfseries \href{https://github.com/faustomilletari/CFCM-2D}{Fork me on GitHub} };
		\draw[forkmefg!60, dashed, line width=.08em, dash pattern=on .5em off 1.5\pgflinewidth] (-25em,1.2em) rectangle (25em,-1.2em);
		\end{tikzpicture}
	};
	\end{tikzpicture}
}

\usepackage{graphicx}
\graphicspath{{pics/}{figs/}}

\begin{document}

\frontmatter          
\pagestyle{headings}  
\mainmatter              

\title{CFCM: Segmentation via Coarse to Fine Context Memory}
\titlerunning{CFCM: Segmentation via Coarse to Fine Context Memory}  %

\author{
 Fausto Milletari\inst{1},
 Nicola Rieke\inst{1,2},
 Maximilian Baust \inst{2, *},
 Marco Esposito \inst{2},
 Nassir Navab \inst{2}
}

\institute{
 NVIDIA \and 
 Technische Universit\"at M\"unchen
}
\authorrunning{F.Milletari, N.Rieke, M. Baust, M. Esposito, N. Navab}

\maketitle              

\forkme[east]

\blfootnote{* Maximilian Baust is now working for Konica Minolta Laboratory Europe}
\begin{abstract}
Recent neural-network-based architectures for image segmentation make extensive usage of feature forwarding mechanisms to integrate information from multiple scales.
Although yielding good results, even deeper architectures and alternative methods for feature fusion at different resolutions have been scarcely investigated for medical applications.
In this work we propose to implement segmentation via an encoder-decoder architecture which differs from any other previously published method since (i) it employs a very deep architecture based on residual learning and (ii) combines features via a convolutional Long Short Term Memory (LSTM), instead of concatenation or summation.
The intuition is that the memory mechanism implemented by LSTMs can better integrate features from different scales through a coarse-to-fine strategy; hence the name Coarse-to-Fine Context Memory (CFCM).
We demonstrate the remarkable advantages of this approach on two datasets: the Montgomery county lung segmentation dataset, and the EndoVis 2015 challenge dataset for surgical instrument segmentation.
\end{abstract}

\section{Introduction and previous work}
\label{sec:intro}
The usefulness of multi-scale feature representations has been acknowledged by the computer vision community for decades, \cf Burt and Adelson \cite{burt1987laplacian} or Koenderink and van Doorn \cite{koenderink1987representation} for instance, and both traditional and modern approaches for image segmentation, registration or stereo rely on the integration of features from multiple scales.
In recent years, convolutional neural networks (CNNs) have advanced the state of art in image segmentation tremendously as this technique allows to learn rich feature representations over multiple scales.
However, the integration of these representations for obtaining full-resolution segmentations is not straightforward and it is an active field of research.

Besides applying CNNs in a patch-based fashion as proposed by \cite{ciresan2012deep}, which is still common practice for very large data such as whole slide images in digital pathology, FCNNs making use of whole-image information, originally suggested by Long \etal \cite{long2015fully}, have turned out to be powerful tools.
Based on this work, Ronneberger \etal \cite{ronneberger2015u} extended the idea of feature forwarding and proposed a symmetrical architecture. In this work the expanding or decoding path takes advantage of fine-grained features from the compressing path that are forwarded via skip connections.
Feature forwarding has also turned out to be a very successful concept for 3D volumetric segmentation as demonstrated by Milletari \etal \cite{milletari2016v} and \c{C}i\c{c}ek \etal \cite{cciccek20163d}.
Further achievements to these architectures have then been accomplished by improved up-sampling, \eg \cite{badrinarayanan2017segnet}, improved training strategies, \eg \cite{hwang2016self}, integration of random fields and \`{a}trous convolutions \cite{chen2016deeplab}, and particularly the application of residual learning \cite{he2016deep}, \eg \cite{chen2017voxresnet,laina2017concurrent}.
For a more complete review of related works, we refer the interested reader to the recent review of Litjens \etal \cite{litjens2017survey}.
In summary, it can be said that most state-of-the-art segmentation architectures use skip connections for feature forwarding and multi-scale context integration.
However, most current approach resort to simple feature fusion schemes, based on concatenation or summation.
An exception is represented by the gated feedback refinement network by Islam \etal \cite{islam2017gated} which comprise gate units to control the information flow and filter out ambiguity.

In this work, we present an alternative approach to multi-scale feature integration based on Long-Short-Term-Memory-units (LSTMs) initially proposed by Hochreiter and Schmidhuber \cite{hochreiter1997long}, wich we term Coarse-to-Fine Context Memory (CFCM).
The rationale behind this approach is that LSTMs implement a memory mechanism in which information can be maintained through different steps and only be updated with new information when necessary. We employ this idea to manage features extracted at different resolutions from the compressing path of the network.
To demonstrate the potential of this approach, we compare our method to established architectures on two different datasets.

\section{Method}
Our segmentation approach is based on a fully convolutional architecture consisting of an encoding and a decoding part, \cf Figure \ref{fig:architecture}.
While encoding is based on a standard ResNet architecture, decoding is implemented using convolutional LSTMs. 
The core idea of this approach is to use a memory mechanism, implemented via convolutional LSTMs, for fusing features extracted from different layers of the encoder.
Thereby, the convolutional LSTMs take the role of a coarse-to-fine focusing mechanism which first perceives the global context of the input data, as the deepest activations are fed to the inputs of the LSTM, and later processes fine-grained details.
This happens when shallower, high-resolution features are considered. Code available on \href{http://github.com/faustomilletari/CFCM-2D}{http://github.com/faustomilletari/CFCM-2D}.

\subsection{Encoder}
Recent works~\cite{ronneberger2015u,milletari2016v,cciccek20163d} have proven that forwarding features extracted by the layers of the encoding path to the corresponding layers of the decoding path greatly improves performance:
At training time, convergence can be achieved within a smaller number of epochs, and at testing time the segmentation performance is better.
To this end, feature fusion strategies based on concatenation and summation have been employed by various authors~\cite{cciccek20163d,laina2017concurrent,long2015fully,ronneberger2015u}, but alternatives have been rarely investigated, which constitutes one of the motivations for this work.
Our aim is to model the hierarchical nature of the features we extract from the encoding path explicitly in order to build a principled and more effective way of fusing them.

\begin{figure*}[t] 	
\centering 	
\includegraphics[width = \textwidth]{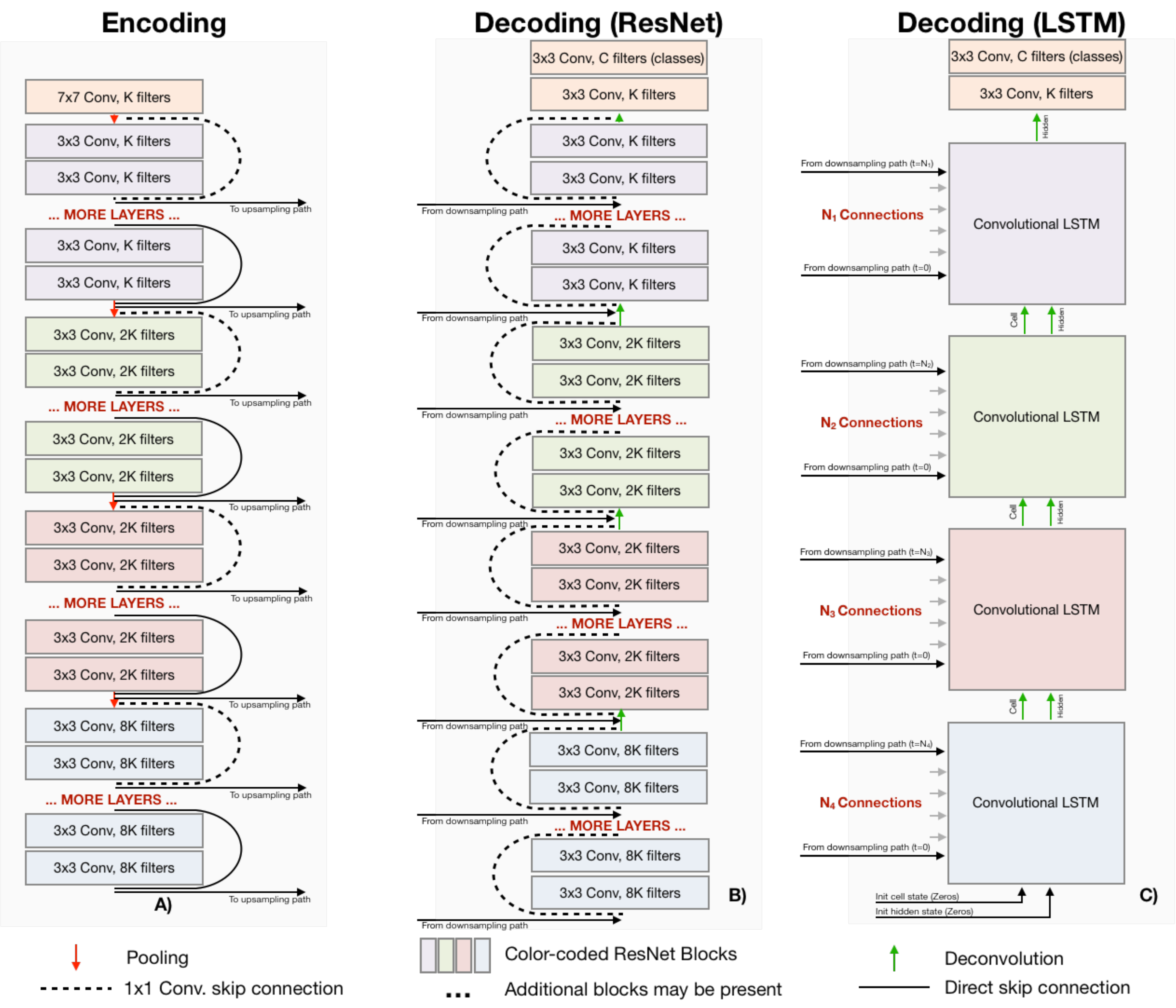} \caption{\textbf{Graphical Representation} of the ResNet+Skip connection architecture and the proposed Coarse to Fine Context Memory (CFCM) based on LSTM. The number of layers in each block of the ResNet varies according to the architecture (ResNet-18, -34, -50, -101). The number of skip connection follows accordingly.} \label{fig:architecture} 
\end{figure*}

As shown in Figure \ref{fig:architecture}, we employ a ResNet architecture and we derive features at each residual block.
These features are interpreted as a coarse-to-fine scale sequence, starting from the bottom of the ResNet up to its top. 
The deepest features are characterized by low resolution but high receptive field.
As shown by Zeiler \etal~\cite{zeiler2014visualizing} as well as other recent works, these features are taking into account global image information and high-level, complex patterns.
Due to their coarse resolution, however, they do not yield information about fine-grained details.
The uppermost features, on the other hand, refer to much more low-level, and fine-grained details, which is due to their high resolution and their limited receptive field.

\subsection{Decoder}

\begin{figure*}[t]	
\centering 	
\includegraphics[scale=0.20]{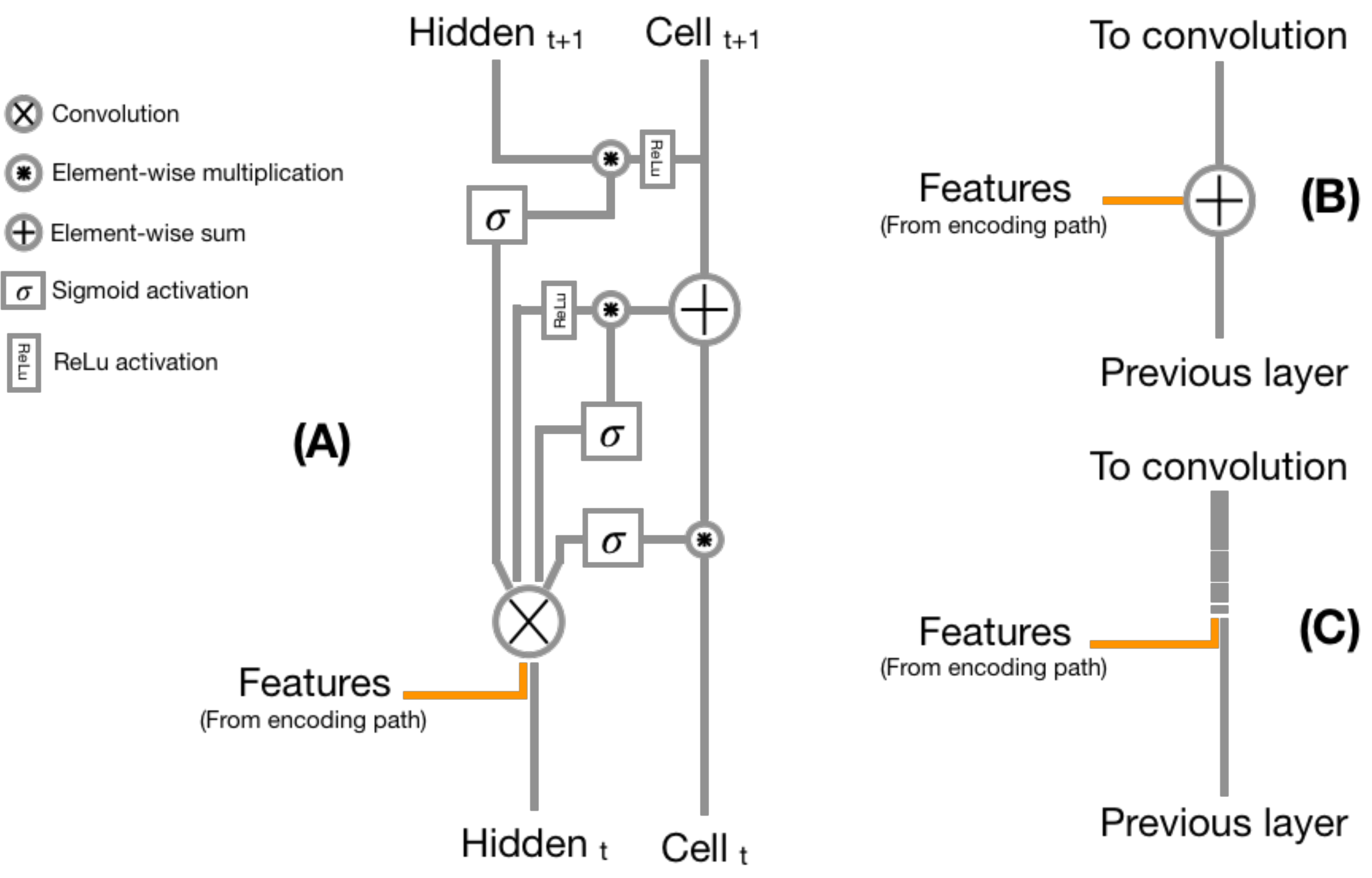} 	
\caption{\textbf{Schematic Comparison} of convolutional LSTMs (A) and feature fusion by summation (B) and concatenation (C).} \label{fig:lstm} 
\end{figure*}

Our decoder treats each block of the ResNet encoder as a single time-step. As shown in Figure \ref{fig:architecture} we forward the outputs of these blocks to our decoder, where the features are processed through LSTM cells.
To this end, we employ convolutional LSTMs \cite{xingjian2015convolutional}, which have the capability of selectively updating their internal states at each step depending on the result of a convolution.
As shown in Figure \ref{fig:lstm} each time step makes use of three feature sets: inputs, hidden and cell state.
Inputs are concatenated with the hidden state.
A convolution is performed and its result is used to (1) pass a part of the information stored in the cell state through the forget gate;
(2) compute new activations which contribute to the cell state after being (3) decimated;
(4) compute a new hidden state.

The initial hidden and cell states of the first LSTM are set to zero.
The states of all other LSTM blocks in the decoder are initialized to be the up-sampled versions of the hidden and cell activations of the cell below.
Intuitively, this mechanism can be understood as a coarse-to-fine context integration mechanism.
The whole context of the picture is perceived first and, gradually, fine-grained details are added as the feature receptive fields decrease. 
Compared to other strategies, depicted also in Figure \ref{fig:lstm}, this architecture allows global context to be kept in memory while details are gradually being added.
This aims at imitating the mechanism allowing humans to focus on details of an object while keeping in mind its global appearance. 
The hidden state of the last LSTM cell of the decoder constitutes the output of the memory mechanism and the last two convolutional layers produce the final segmentation. 

\section{Experimental evaluation}
To demonstrate the advantages and general applicability of the proposed technique, we evaluate the segmentation performance on two different datasets.
First, we compare CFCM against to two baselines, U-Net and ResNet+Skip, on the Montgomery Country X-ray Dataset.
In the second experiment, we focus on the general applicability and test the performance regarding segmentation of surgical instruments in endoscopic surgery sequences.
To show the superior performance of the method, we compare to state-of-the-art networks that are specialized on instrument tracking. 

\textbf{Implementation details}
Our networks are initialized with the same set of parameters, trained with batch-size $16$ with a learning rate of $0.00001$ optimizing for the dice coefficient \cite{milletari2016v}. When learning to segment the Montgomery county X-Ray we train for 150 epochs. When dealing with EndoVis we train for 30 epochs. The images are scaled to an input size of $256\times256$ pixels. Our ResNet+Skip connection is depicted in Figure \ref{fig:architecture}. Its decoder is a mirrored version of the encoder with skip connections at each block.
Our method CFCM is implemented in tensorflow and will be made publicly available upon paper publication. 

\begin{figure} 	
\centering 	
\includegraphics[width = \textwidth]{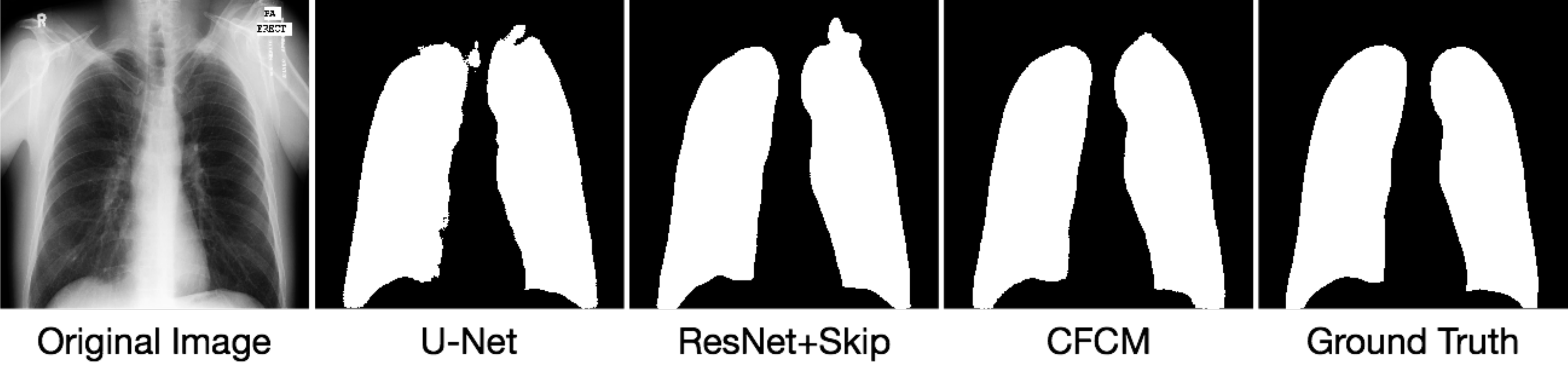} 	
\caption{Qualitative results on the Montgomery county X-Ray dataset.} \label{fig:qualitative_xray} 
\end{figure}

\textbf{Montgomery County X-ray Set.}
\label{sec:exp:MC}
This dataset comprises 138 annotated posterior-anterior chest x-rays and has been acquired from the tuberculosis control program of the Department of Health and Human Services of Montgomery County, MD, USA \cite{jaeger2014two}.
The set contains 80 normal cases and 58 abnormal cases with manifestations of tuberculosis including effusions and miliary patterns.
For testing, we perform a three fold cross evaluation for binary lung segmentation and report the mean scores in Table~\ref{tab:MC}.
As shown in Figure~\ref{fig:qualitative_xray}, U-Net and ResNet tend to misclassify the air-filled upper trachea or fractions of the shoulder as part of the lung, while the proposed CFCM is successful in capturing the global shape and fine outlines of the anatomy. 
Especially the leakage to the region of the shoulder is reduced.
The improved performance is also reflected in consistent better quantitative results (see Table~\ref{tab:MC}).
It can be observed that the performance of CFCM improves with depth while the ResNet with simple skip connections starts to overfit due to high number of parameters.

\begin{table}[]
    \centering
    \caption{\textbf{Results for Montgomery County X-Ray Set}. \newline \hspace{\textwidth} Abbreviations: \textit{DICE} = Dice coefficient, \textit{MAD} = Mean Absolute Distance, \textit{RMS} = Root-Mean-Square distance, \textit{HD} = Hausdorff Distance}
\resizebox{\columnwidth}{!}{
\begin{tabular}{  l  c | c | c| c }
\hline
Architecture & DICE & MAD & RMS & HD \\ 
\hline
U-Net & $0.961\pm0.020$ & $0.160\pm0.288$ & $0.960\pm\mathbf{1.207}$ & $20.463\pm15.607$\\
\hline
ResNet18+Skip & $0.966\pm0.0239$ & $0.228\pm0.475$ & $1.265\pm1.865$ & $21.670\pm19.529$\\
ResNet34+Skip & $0.969\pm0.0217$ & $0.189\pm0.400$ & $1.034\pm1.664$ & $16.335\pm14.959$\\
ResNet50+Skip & $0.969\pm0.0225$ & $0.201\pm0.456$ & $1.072\pm1.809$ & $17.015\pm16.651$\\
ResNet101+Skip & $0.969\pm0.0222$ & $0.198\pm0.414$ & $1.120\pm1.766$ & $17.340\pm16.807$\\
\hline
CFCM18 & $0.967\pm0.019$ & $0.151\pm0.298$ & $0.907\pm1.324$ & $15.838\pm14.853$\\
CFCM34 & $0.969\pm0.0189$ & $0.145\pm0.328$ & $0.844\pm1.414$ & $14.984\pm15.078$\\
CFCM50 & $0.970\pm0.0188$ & $0.144\pm0.311$ & $0.863\pm1.414$ & $14.848\pm15.151$\\
CFCM101 & $\mathbf{0.972}\pm\mathbf{0.0181}$ & $\mathbf{0.143}\pm\mathbf{0.307}$ & $\mathbf{0.821}\pm1.379$ & $\mathbf{14.238}\pm\mathbf{15.011}$\\
\hline
\end{tabular}
}
    \label{tab:MC}
\end{table}


\textbf{EndoVis 2015\footnote{MICCAI 2015 Endoscopic Vision Challenge Instrument Segmentation and Tracking Sub-challenge \url{http://endovissub-instrument.grand-challenge.org}}.}
\label{sec:exp:EndoVis}
The dataset covers in total $6$ \emph{ex-vivo} endoscopic surgery sequences of image resolution $720 \times 576$ pixels.
The training data contains four 45s sequences.
The remaining 15s of the same sequence together with two new 60s videos form the testing dataset.
\begin{figure} 	
\centering 	
\includegraphics[width = \textwidth]{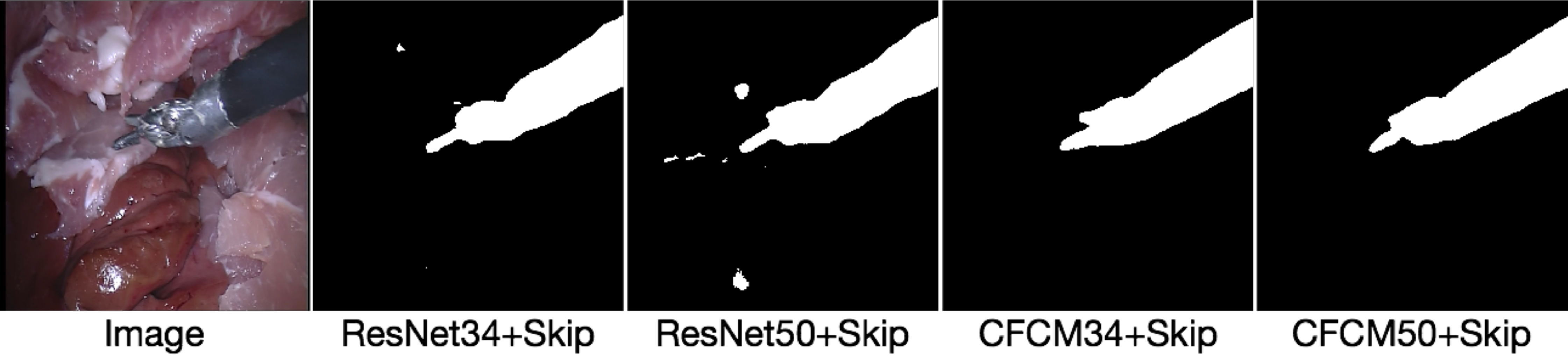} 	
\caption{Qualitative results on the EndoVis dataset.} \label{fig:qualitative_EndoVis} 
\end{figure}
There are three semantic classes (manipulator, shaft and background).
As specified in the guidelines, we preformed a cross-validation by leaving one surgery out of the training data.
The segmentation result was compared to generic methods as well as algorithms that were explicitly published for this task~\cite{laina2017concurrent,garcia2016real,pakhomov2017deep}.
Garcia P. Herrera \emph{et al.} ~\cite{garcia2016real} proposed a Fully convolutional network for segmentation in minimally invasive surgery.
To achieve real-time performance, they applied the network only on every couple of frames and propagated the information with optical flow (FCN+OF).
DLR~\cite{pakhomov2017deep} represents a deep residual network with dilated convolutions.
Laina and Rieke \emph{et al.}~\cite{laina2017concurrent} suggested a unified deep learning approach for simultaneous segmentation and 2D pose estimation using Fully Convolutional Residual Network with skip connections. 
As depicted in Table~\ref{tab:EndoVis}, we outperform state of the art for both binary segmentation as well as multi-class segmentation.
The major advantage of the proposed method over alternative approaches can be seen in the robustness to specular noise and the precision for the grasper (Table~\ref{tab:EndoVis}, Figure~\ref{fig:qualitative_EndoVis}).
While the other methods have problems with the most flexible part of the instrument, CFCM can still recover the fine segmentation by the deep feature integration with LSTMs.

\begin{table}[]
    \centering
    \caption{\textbf{Results for EndoVis} \newline \hspace{\textwidth} Abbreviations: \textit{B.Acc} = Balanced Accuracy, \textit{Rec} =  Recall, \textit{Spec} = Specificity, \textit{DICE} = Dice coefficient}
\resizebox{\columnwidth}{!}{
 \begin{tabular}{  l r  r  r  r | r  r  r  r |  r r r r}
      \hline
     & \multicolumn{4}{c}{Binary}  &  \multicolumn{4}{c}{Shaft} &   \multicolumn{4}{c}{Grasper} \\
Method   & \multicolumn{2}{r}{B.Acc.}     Rec. &  Spec. &  DICE & Prec. & Rec. &  Spec. & DICE & Prec. & Rec. &  Spec. & DICE \\ \hline
   FCN~\cite{garcia2016real} & $83.7$ & $72.2$ & $95.2$ & - & - & - & - &- &- & - &- &- \\
   FCN+OF~\cite{garcia2016real} & $88.3$ & $87.8$ & $88.7$ & - & - & - &- &- &- &- & - & - \\ 
   DRL~\cite{pakhomov2017deep} & $92.3$ & $85.7$ & $98.8$ & - & - & - &- &- &- &- & - & -  \\ 
   CSL~\cite{laina2017concurrent} & $92.6$ & $86.2$	 & $\mathbf{99.0}$ &  $88.9$ & $92.4$ & $\mathbf{83.8}$ & $\mathbf{99.3}$ & $\mathbf{87.7}$ & $79.1$ & $77.4$ & $99.2$ & $77.7$ \\ 
   \hline
   ResNet18+Skip & $97.3$ & $86.0$ & $98.9$ & $87.5$ & $97.5$ & $82.7$ & $98.8$ & $83.5$ & $98.4$ & $70.2$ & $99.3$ & $73.1$  \\ 
   ResNet34+Skip & $97.4$ & $84.9$ & $98.9$ & $87.2$ & $83.6$ & $82.4$ & $98.9$ & $83.0$ & $\mathbf{98.6}$ & $74.2$ & $\mathbf{99.4}$ & $75.8$  \\
   ResNet50+Skip & $97.2$ & $85.3$ & $98.8$ & $86.8$ & $97.5$ & $82.0$ & $97.6$ & $83.5$ & $98.5$ & $75.9$ & $99.3$ & $76.0$  \\ 
   ResNet101+Skip & $97.0$ & $84.9$ & $98.7$ & $86.0$ & - & - & - & - & - & - & - & -  \\ 
   \hline
   CFCM18 & $97.7$ & $88.0$ & $98.9$ & $89.3$ & $97.4$ & $81.3$ & $99.1$ & $81.5$ & $98.3$ & $73.6$ & $99.2$ & $75.0$ \\
   CFCM34 & $\mathbf{97.8}$ & $\mathbf{88.8}$ & $98.8$ & $\mathbf{89.5}$ & $\mathbf{97.6}$ & $82.1$ & $99.1$ & $84.0$ & $\mathbf{98.6}$ & $77.0$ & $99.3$ & $\mathbf{78.3}$  \\ 
   CFCM50 & $97.5$ & $87.7$ & $98.8$ & $88.8$ & $\mathbf{97.6}$ & $81.4$ & $99.0$ & $83.9$ & $98.5$ & $\mathbf{77.8}$ & $99.2$ & $78.1$  \\ 
   CFCM101 & $97.2$ & $81.1$ & $\mathbf{99.0}$ & $85.5$ & - & - & - & - & - & - & - & -  \\ 
   \hline
\end{tabular}
}
    \label{tab:EndoVis}
\end{table}

\section{Conclusion}
We presented a novel approach for CNN-based image segmentation that achieves multi-scale feature integration via LSTMs which we term Coarse-to-Fine Context Memory (CFCM).
This approach has been evaluated on two challenging segmentation databases of chest radiographs and video data showing surgical instruments during an intervention in endoscopy.
The experiments demonstrate that the proposed method achieves superior performance and can outperform generic as well as application-specific networks.
Future research might include the extension of this concept to 3D and the exploration of different memory mechanisms.

\bibliographystyle{splncs03}
\bibliography{bibliography}

\end{document}